\documentclass{article}

\usepackage{appendix}

\usepackage{PRIMEarxiv}

\usepackage{amsmath}

\usepackage[utf8]{inputenc} 
\usepackage[T1]{fontenc}    
\usepackage{hyperref}       
\usepackage{url}            
\usepackage{booktabs}       
\usepackage{amsfonts}       
\usepackage{nicefrac}       
\usepackage{microtype}      
\usepackage{lipsum}
\usepackage{fancyhdr}       
\usepackage{graphicx}       
\graphicspath{{media/}}     

\title{TREB: a BERT attempt for imputing tabular data imputation
}

\author{
  Shuyue Wang \\
  Shanghai A\&I Co., Ltd. \\
  Shanghai\\
  \texttt{henri\_w\_91@hotmail.com}\\  
  \And
  Wenjun Zhou \\
  Huawei Technologies Co., Ltd. \\
  Shanghai \\
  \texttt{2260882790@qq.com}\\  
  \And
  Han \textit{drk-m-s} Jiang \\
  M3, Autoslide Inc.,Talent Mediocre Holdings Group,  \\
  Paris, Grenoble, Talinn\\
  \texttt{john.jianghan@gmail.com}\\
  \And
  Shuo Wang \\
  Digital Medicial Research Center, School of Basic Medical Sciences, Fudan University \\
  Shanghai \\
  \texttt{shuowang@fudan.edu.cn} \\
   \And
  Ren Zheng \\
  Shanghai A\&I Co., Ltd. \\
  Shanghai \\
  \texttt{zhengren@jingzhi-sh.com}\\
}

\begin{document}
\maketitle

\begin{abstract}
TREB, a novel tabular imputation framework utilizing BERT, introduces a groundbreaking approach for handling missing values in tabular data. 
Unlike traditional methods that often overlook the specific demands of imputation, TREB leverages the robust capabilities of BERT to address this critical task.
While many BERT-based approaches for tabular data have emerged, they frequently under-utilize the language model's full potential. 
To rectify this, TREB employs a BERT-based model fine-tuned specifically for the task of imputing real-valued continuous numbers in tabular datasets.
The paper comprehensively addresses the unique challenges posed by tabular data imputation, emphasizing the importance of context-based interconnections.
The effectiveness of TREB is validated through rigorous evaluation using the California Housing dataset. 
The results demonstrate its ability to preserve feature interrelationships and accurately impute missing values.
Moreover, the authors shed light on the computational efficiency and environmental impact of TREB, quantifying the floating-point operations (FLOPs) and carbon footprint associated with its training and deployment.
\end{abstract}

\keywords{tabular imputing \and BERT  \and tokenizer \and RoBERTa \and TREB}

\section{Introduction}
Data in the real world often manifests in a form incorporating tabular elements.
Tabular data, a stalwart of structured information for decades, is organized into mazes of rows and columns.
This ubiquitous format has found applications in a multitude of fields, ranging from finance and medicine to business, agriculture, education, and beyond.
These attributes can depend on one another and have different data types (i.e., categorical and numerical data).
Fang et al.\cite{fang2024large} summarizes characteristics of tabular data, the last three of which constitutes the driving force of the proposed approach in this paper\label{lopk}: heterogeneity, sparsity, lack of prior knowledge, dependency on pre-processing, context-based interconnection, and order invariant.
Traditionally, tabular data has been used for simply organizing and reporting information. 
Over the past decade, its usage has evolved significantly due to machine learning competitions and open-source contributions.\cite{towardsai2023};

A significant obstacle for general tabular data problems is the relatively limited size of available datasets. This contrasts sharply with domains like vision and NLP, where vast amounts of "extra" data can be readily sourced from the internet. 
For instance, learning to recognize an image of a cat involves studying a seemingly endless collection of feline images.
This task is ubiquitous in computer vision, explicitly manifested in input (e.g., user prompts), processing (e.g., attention-based heatmaps), or output (e.g., generated cat images or captions). 
The universality of the concept of a 'cat' further underscores its prevalence.
Similar observation can also be made in the field of NLP problem in that every task is explicit and universal, whilst tabular data is independent from table to table because they represent  different observation of different things in the real world, let alone the discrepancy and biases in observation. 
In this case, titles of tables seems to be of help; however, they  lack in the universality  compared against the above examples, which enhances the difficulty in finding a proper solution.
This aligns with the notion of a "lack of prior knowledge" (cf. \ref{lopk}). 
In contrast, models like CLIP, BLIP, or GLIP can effectively assist diffusion models in computer vision tasks.

Before the rise of AI, people relied on traditional table perception platforms, such as manual spreadsheets, SQL databases, statistical software, programming libraries, and visualization tools. 
Even simple techniques like SMOTE interpolation (\cite{chawla2002smote}) can outperform complex tabular GANs (\cite{camino2020oversampling}) for minority class oversampling.\cite{grinsztajn2022tree}
Tree-based ensemble methods, especially gradient-boosted decision trees, remain the state-of-the-art for tabular data predictions (\cite{gorishniy2021revisiting}, \cite{chen2016xgboost}).
Efforts to integrate neural networks with tabular data for management tasks are ongoing (\cite{fang2024large}). Additionally, borrowing approaches from other domains, such as leveraging tree-based structural knowledge (\cite{ke2018tabnn}) or topological information via directed graphs (\cite{ivanov2021boost}), are common techniques.
GBDT is one such strong and lightweight technique in modelling tabular data, which powers some implementations of non-deep-learning neural network in leveraging expressive feature combinations.\cite{ke2018tabnn}

Artificial intelligence is often considered to be essentially equivalent to modern statistics, albeit with a different terminology (\cite{Sargent2023})\footnote{Thomas J. Sargent, laureate of Nobel Economics Prize in 2011, said  Artificial intelligence is just statistics., on August 11th, 2018, in Global Science and Technology Innovation Forum (GSTIF)}. 
While statistics draw inspiration from various fields like physics, biology, and economics, it remains a valuable tool, especially when reasoning fails. 
In fact, statistics can be viewed as a last resort (\cite{chen2021}) in many situations.\footnote{Thanks to Baojun He for identifying the quotation source.}
Consider a table where relationships between columns are known. Some columns may be directly influenced by others, while others exhibit more complex, non-linear relationships. 
Traditional statistics often rely on correlation metrics to quantify these relationships, but they may not provide an explicable metrics of correlation. 
It is in such cases that generative AI is sought to offer the potential to delve deeper into the underlying dynamics of tabular data manipulation.

Generative AI, or GenAI, reached a significant milestone with the emergence of ChatGPT in late 2022. 
The successful implementation of attention mechanisms demonstrated that 'thinking' could be achieved by strategically accumulating and utilizing information for program execution, challenging traditional research paradigms.
Despite significant advancements in the past three years, it's clear that foundation models are not a direct representation of the real world. 
They are intelligent understanding machines. 
Tabular data researchers rarely expect to simply call a foundation model's API for direct table imputation.
Even when prompt engineering (\cite{wang2023brief}) and increased input token capacity (\cite{qin2024mooncake}) are considered, the results are often unsatisfactory. 
This highlights the need for further research to bridge the gap between foundation models and specific tabular data tasks.

\begin{figure}[htbp]
    \centering
    \includegraphics[width=0.99\textwidth]{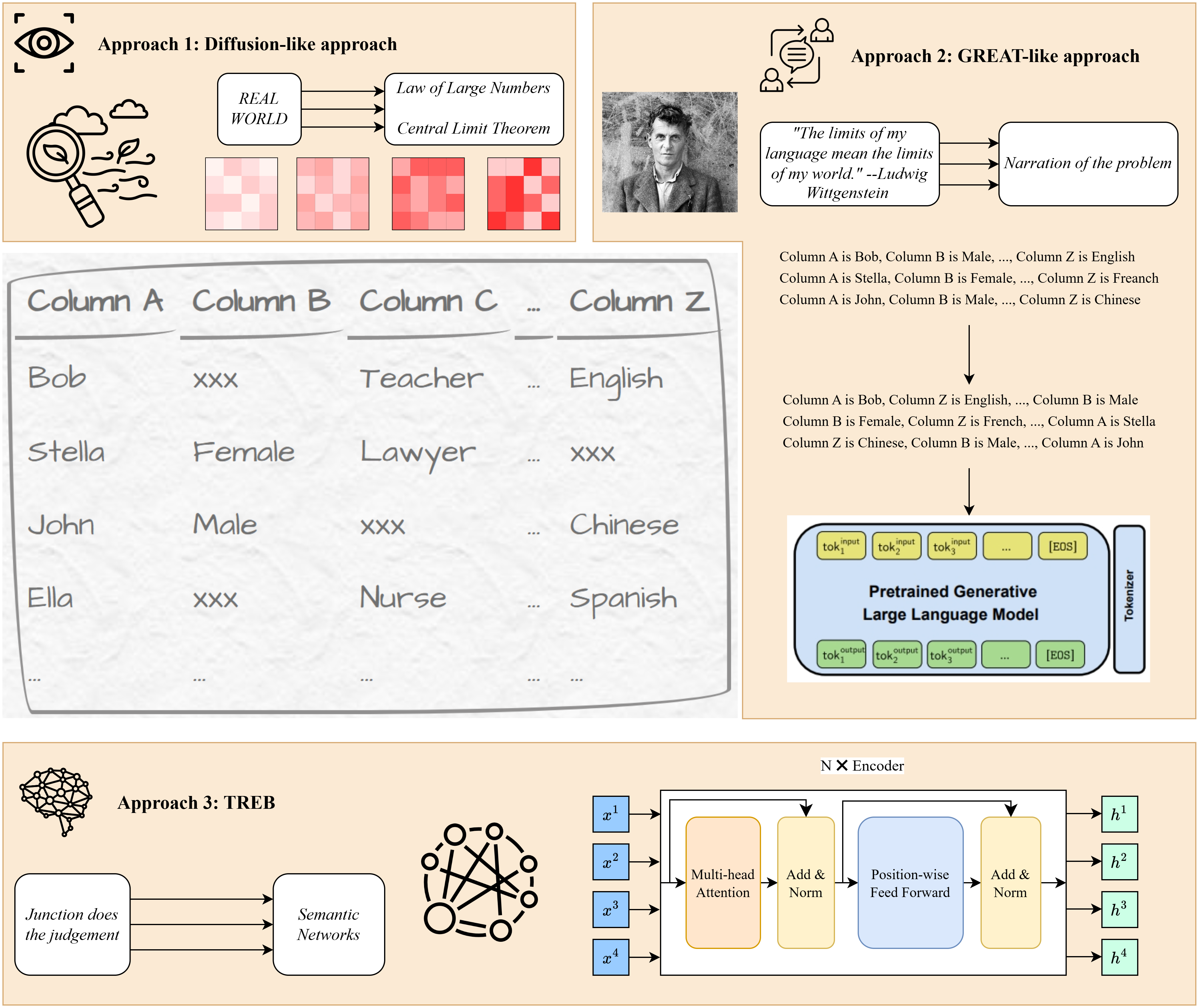}
    \caption{The three major types summarized in GenAI approaches in tabular data manipulation: 1. using diffusion model; 2 .using pretrained language model (e.g. GPT2); 3. using BERT (to which this paper belongs).}
    \label{fig-1-carcasse}
\end{figure}

Based on a preliminary investigation, this paper tentatively categorizes generative AI approaches in the realm of tabular data manipulation into three primary paradigms, as illustrated in fig.\ref{fig-1-carcasse}.
The first approach is based on intuitive understanding. 
Assuming underlying mechanisms are beyond human perception, we can leverage the laws of large numbers and the central limit theorem to obtain macro-scale observations. 
By distilling knowledge from these observations, we can sample the data.
A recent trend of such genre in this area involves extending generative AI models and their variants to general tabular problems. 
TabDDPM (\cite{kotelnikov2023tabddpm}) demonstrates the effectiveness of diffusion models for tabular data with heterogeneous features. 
CTGAN (\cite{xu2019modeling}) uses a conditional generator to address the challenges of modeling row probability distributions and generating realistic synthetic data.
These diffusion model approaches often require transforming heterogeneous tabular input into homogeneous data (\cite{fang2024large}).
The second approach involves language model, transforming tabular data into a language-like format, reflecting the idea that language is a primary carrier of knowledge: \textit{The limits of my knowlede mean the limits of my world.}\\cite{wittgenstein2023tractatus}.
Recent advancements in large language models have enabled various tasks related to tabular data modeling, including table understanding and data generation (\cite{hegselmann2023tabllm}, \cite{sui2023tap4llm}).
The approach usually relies on the semantic knowledge of LMs to transfer feature patterns from pre-training feature names to downstream ones, implicitly requiring meaningful and clear feature semantics.\cite{yan2024making}
Borisov et al. proposed a comprehensive linguistic description approach, transforming tabular data while considering data types, ranges, continuous/categorical attributes, and collation methods (\cite{borisov2022language}). 
Another less intrusive method involves prompt engineering (\cite{wang2023brief}), chain-of-thought agents (\cite{yang2023effective}), and retrieval-augmented generation (\cite{sundar2023ctbls}) to leverage the power of language models for tabular data problems.
The third approach is what the proposed approach in this paper belongs to.
It is classified as 'attention-based methods' by Fang et al.\cite{fang2024large}, where basically BERT  (Bidirectional Encoder Representations from Transformers) is dominating the role.\cite{koroteev2021bert}
The application of BERT in tabular data generation has gained traction as researchers explore ways to leverage its capabilities for improving synthetic data generation processes.
As matter of fact, TREB is a \textit{verlan} of BERT, indicating the method's nature amongst all tabular data tools.
Detailed description will be in sections\ref{sec:related_works} and \ref{sec:treb}.

In the realm of generative AI research concerning tabular data, the term 'data manipulation' often encapsulates the process of 'sampling,' although 'imputation' is also occasionally included. 
While synthesizing tabular data entails the creation of an artificial table (or multiple relational tables) based on an original dataset, imputation represents a form of 'conditional sampling'. 
Broadly speaking, sampling adopts a macro-scale perspective, while imputation is a more focused approach.
Imputation serves as a valuable tool for data augmentation, particularly effective in addressing class imbalances. 
Its primary objective is to preserve and leverage the interrelationships between features represented by columns, thereby accurately and precisely filling missing values. 
In contrast, sampling seeks to simulate a broad-scale representation of data distribution patterns within a limited number of rows.
This paper concentrates on developing a BERT-based approach for imputing tables containing real continuous numbers, thereby introducing the TREB (TabluaR imputatEr using Bert) method.

In this paper, section \ref{sec:related_works} reviews the past literature of Bert architecture in tabular data studies; section \ref{sec:treb} provides the mathematical description of the proposed approach and illustrates the fundamental steps it takes to accomplish the task in discussion, where subsection \ref{sec:exp} demonstrates the performance in imputing based on California Housing data set\cite{pace1997sparse}, where accuracy is the priority in the evaluation. 
The cost of computation in terms of flops and carbon footprint is also shown there. 
The appendix includes some figures.
The code repository is made public for your convenience\footnote{\url{https://github.com/shuyueW1991/TREB}}.

\section{Related works} \label{sec:related_works}
\paragraph{BERT and tabular data}
BERT (Bidirectional Encoder Representations from Transformers) is a deep learning model designed to pre-train representations from unlabeled text data \cite{kenton2019bert}. 
Like all transformer-related approaches, BERT is tailored for problems where both the input and output information are sequences. 
A sequence is an arbitrary set of contiguous text tokens \cite{graves2012sequence}.
BERT addresses a fundamental challenge in Natural Language Processing (NLP): text representation. 
Text representation involves transforming natural language data into a format interpretable by machines. 
While simple encodings exist, effective NLP tasks require representations that capture the inherent meaning and structure of the text \cite{torfi2020natural}.
BERT's innovation lies in its unique text representation methodology and training paradigm. 
Its training comprises two distinct phases: unsupervised pre-training on a large corpus of unlabeled text and subsequent fine-tuning on a labeled dataset tailored to a specific downstream task \cite{vaswani2017attention}. 
While the fine-tuning process and architectural modifications may vary depending on the task, the underlying model and parameter set remain consistent.
Although BERT's autoencoder architecture eliminates the need for task-specific adaptation, it introduces artificial tokens like \verb|[SEP]| and \verb|[MASK]| during pre-training. 
This discrepancy between the pre-training and fine-tuning environments can potentially hinder the model's performance on downstream tasks \cite{koroteev2021bert}. 
Moreover, BERT's initial training on general-purpose text may not perfectly align with the data distribution of target application domains. 
To address this, fine-tuning or even retraining can be conducted on different tasks, subjects, and datasets, contingent upon the nature of the text data and available resources\cite{sun2019fine}.

\paragraph{TABERT}
While numerous BERT-based studies have explored various aspects of tabular data, investigations specifically targeting imputation tasks remain notably scarce. To provide a contextual understanding of how BERT is applied to tabular data, here we present a brief overview of the TABERT model.\cite{yin2020tabert}
As one application of BERT in tabular data, it  is trained on a substantial corpus of 26 million tables and their corresponding English contexts. 
Built upon the BERT architecture, TABERT effectively acquires contextual representations for both utterance tokens (individual words or phrases) and the structured schema of database tables. 
To ensure compatibility with the Transformer-based BERT model, TABERT linearizes the tabular structure. 
This essentially transforms the table into a sequence of elements. 
By utilizing the special token \verb|[SEP]|, TABERT is able to learn representations of utterances and table schemas effectively. 
An illustrative example from the WIKITABLE-QUESTIONS dataset\footnote{\url{stanford.io/38iZ8Pf}} demonstrates this capability.

\section{TREB:  TabluaR imputatEr using Bert.} \label{sec:treb}
\begin{figure}[htbp]
    \centering
    \includegraphics[width=0.99\textwidth]{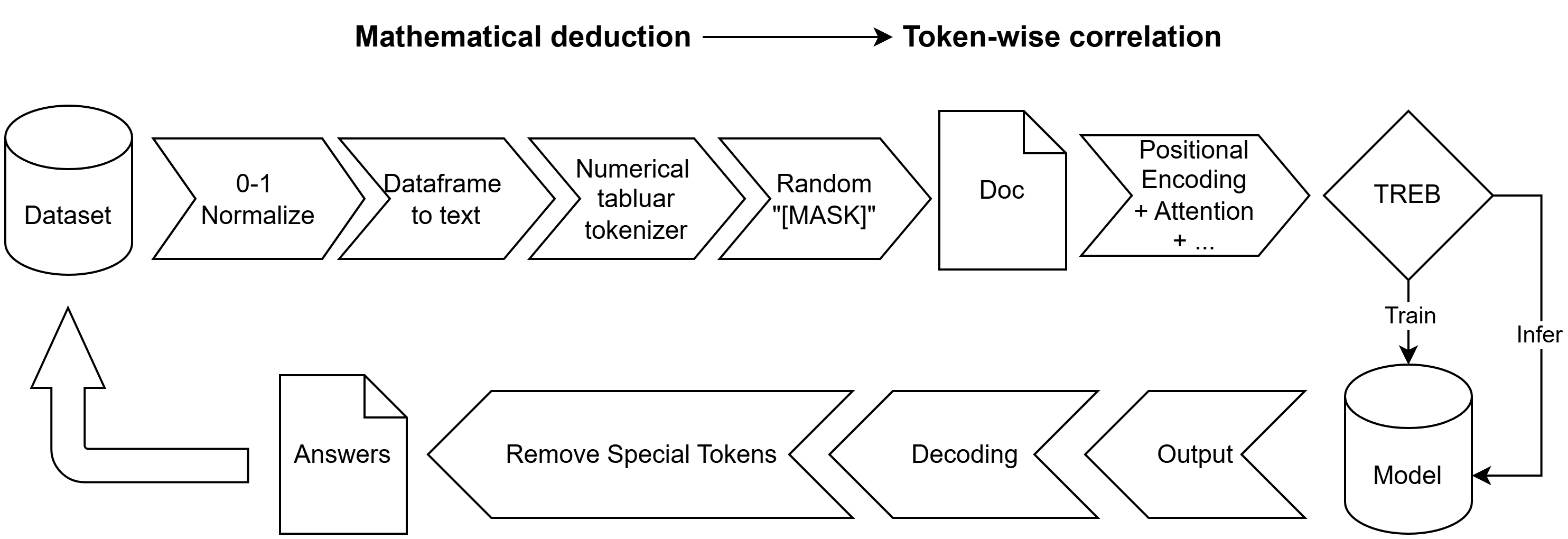}
    \caption{Workflow of TREB.}
    \label{fig:trebworkflow}
\end{figure}

Some authors of BERT studies in tabular data refer schema as the set of columns in a table, and its representation as the list of vectors that represent its columns.\cite{yin2020tabert}
This paper is stick to column-row-value nomenclature, inheriting what is done by Borisov et al.\cite{borisov2022language}
Fig.\ref{fig:trebworkflow} shows the workflow of the proposed approach.

\subsection{Data and tokenization}
To evaluate the effectiveness of our proposed tabular imputation method, we selected the California Housing dataset from Pace et al.\cite{pace1997sparse}
This dataset, characterized by its rich feature set and substantial sample size, aligns well with the requirements of our research. Its numerical nature, devoid of categorical variables, presents an ideal test case for assessing the performance of algorithms specifically designed for tabular data imputation.
To ensure a representative training and validation split, we randomly shuffled the dataset while maintaining stratified sampling to preserve the original distribution across both subsets. 
Consequently, 18,000 rows were allocated to the training set, with the remaining data designated for validation. 
Using the newly defined tokenizer, we established a maximum sequence length of 512. This resulted in a tensor of dimensions (18,000, 512). 
To simulate real-world scenarios, we randomly masked 15
The modified tensor was then integrated into a PyTorch Dataset. Given the computational capabilities of the GPU, we set the batch size to 64 and enabled shuffling for enhanced generalization.

First, we need to convert a row from a table into a string, with the columns names.
For instance, a row is like the one shown in table\ref{tab:sampledrow}.
We do a zero-one normalization, which is basically put every value in a column as: $x_{\text{norm}} = \frac{x - \text{min}}{\text{max} - \text{min}} $, where integer part will be no longer a concern.
Then round it with 4 digits in decimal part, since we would like a game that is only participated by 4-digit tokens, treating the numerical bins as new words.\cite{yan2024making}
We would like the transformer to be the 'interpolator' among limited tokens that has profound numerical meanings.
This is our way of possessing the sensitivity to numeric features seen in other BERT mission in tabular data problems.\cite{yan2024making}
Consequently, it may appear counter-intuitive to employ a vocabulary composed solely of fixed-length digits for these embeddings. 
However, this approach is adopted to facilitate a transition from implicit mathematical reasoning among tabular data towards a token-based game representing numbers, rather than the mathematical deduction based on numbers.\footnote{The decimal part reminds us of the myth question of 'Is 9.9 bigger than 9.11' where many LLM fails to give correct answer. The first author believes it is related to the prior impact given by the tokenization stuff.}

\begin{table}
 \caption{A sampled row from vanilla California Housing Dataset}
  \centering
  \begin{tabular}{ccccccccc}
    \toprule
    \midrule
    MedInc  & HouseAge &	AveRooms&	AveBedrms&	Population&	AveOccup	&Latitude	&Longitude	& MedHouseVal \\
    ...     & ...  &	...&	...      &	...      &	...	     &...	    &...	  & ...    \\
    3.8462  & 52.0 & 	6.281853 & 	1.081081 & 	565.0	 & 2.181467 &	37.85 & -122.25 &3.422 \\
        ... & ...  &	...&	...&	...&	...	&...	&...	& ...    \\
    \bottomrule
  \end{tabular}
  \label{tab:sampledrow}
\end{table}

Next, we make each row a text string in form shown in table\ref{tab:sampledtext}.
Permutation from feature to feature\cite{borisov2022language} is not necessary because we're not dealing with auto-regressive model \cite{xu2024llms}
Since it is now a problem in the realm of language, tools like language model is now available to us.
We lift feature name thus canceling the nuanced manipulation of feature-wise and intra-feature troubles, like the cobbling of 'intra feature attention'\cite{yan2024making}.
In our case, it is BERT tabular imputer, i.e. TREB, evidently.

\begin{table}
 \caption{A sampled text transformed from a row}
  \centering
  \begin{tabular}{l c}
    \toprule
    \midrule
    index  & text \\
    ...     & ...    \\
    314  & Column 0: 0.123, Column 1: 0.4321, .... \\
        ... & ... \\
    \bottomrule
  \end{tabular}
  \label{tab:sampledtext}
\end{table}

For that purpose, we  also need a token that tokenize a string.
Within the BERT architecture, embeddings of plain text, encompassing both numerical and textual elements, encapsulate the semantic properties of words. 
By transforming a discrete set of tokens into a continuous space, information pertaining to words is compressed, and the extensive vocabulary is represented within a lower-dimensional space. 
Faced with 4-digit number in a sentence that is just transformed into a text string:
\begin{quote}
    \begin{verbatim}
        'column 0: 0.2349, column 1: ..., 0.3788']
    \end{verbatim}
\end{quote}
, a regular tokenizer will give you result like this:
\begin{quote}
    \begin{verbatim}
        ['column', '0', ':', '0', '.', '234', '##9',',',
            'column', '1', ':', ...,
                ...,
                    ..., '0', ',', '37', '##8', '##8']
    \end{verbatim}
\end{quote}
where the decimal part is tore up, hindering the game of 4-digit tokens (which is exactly why we set the digit occupation being 4).
It is also observed the pattern shown in all tokenization result does not involve fixed length in terms of decimal parts of a float point number.
The BERT model employs a base tokenizer 'bert-base-uncased' to preprocess textual data. 
This tokenizer serves as a foundational component within the BERT architecture, responsible for transforming raw text into a numerical representation that the model can comprehend and process effectively. 
The tokenizer achieves this by segmenting the text into individual units, known as tokens, which can be either words or subwords. 
Each distinct token is assigned a unique numerical identifier, constructing a vocabulary that enables the model to represent text as a sequence of numbers. 
To facilitate the training process, we establish the 'vocab\_file' as a vocabulary book consisting solely of four-digit numbers, including those prefixed with zeros.\footnote{The 'bert-base-uncase' belongs to AutoTokenizer family that is a little too old that it only accepts add\_tokens for extra addition of vocabulary.}
Within this vocabulary, the token '103' is specifically reserved to represent the '[mask]' token, a crucial element for future training.
The new tokenizer will give us a tokens series like this:

\begin{quote}
    \begin{verbatim}
        ['column', '0', ':', '0', '.', '2349', ',',
            'column', '1', ':', ...,
                ..., 
                ..., '0', '.', '3788']
    \end{verbatim}
\end{quote}
in triple forms input\_ids, attention\_masks and labels.

\subsection{Training paradigm}
We proceeded to the training phase, employing a Roberta model configured with the following parameters: a vocabulary size of 51,100, a maximum sequence length of 514, a hidden size of 768, 12 attention heads, 6 hidden layers, and a type vocabulary size of 1. 
Like many BERT projects in tabular data problems, this paper is built on the basis of RoBERT.\cite{liu2019roberta}
This model was initialized for masked language modeling, tasked with predicting missing words in a given sequence.
Its architecture can be seen in fig.\ref{fig:tRoberta_archi} in Appendix.
An initial learning rate of 1e-4 was adopted. 
A random (15\%) sample of the tokens in the input sequence is selected and replaced with the special token [MASK].
To leverage GPU acceleration, the model and batch were transferred to a GPU device. 
Input IDs, attention masks, and labels were provided to the model, and the outputs were evaluated in terms of token differences to compute the loss. 
Position encoding is lifted as attempts in many other literature\cite{ye2024towards}, but we reserve it here because the column, though incoherent, is inherent in tabular information.
Training was terminated after 500 epochs, at which point the loss converged to approximately 5e-7.

\subsection{Evaluation} \label{sec:exp}
\subsection{Hardware}
We use A800 SXM4 80 GB, which is a professional graphics card by NVIDIA, launched on August 11th, 2022\footnote{\url{https://www.techpowerup.com/gpu-specs/a800-sxm4-80-gb.c3966}}.
The GPU is operating at a frequency of 1155 MHz, which can be boosted up to 1410 MHz, memory is running at 1593 MHz.
The system utilized in this analysis is equipped with a CUDA Driver Version 12.1, specifically driver version 525.60.13. 
The programming language employed is Python, version 3.11.8.final.0 (64-bit). 
The CPU is an Intel Xeon Platinum 8358P with a base clock speed of 2.6000 GHz.

\subsection{Performance}
\subsubsection{Errors}
To evaluate the model's performance to missing data, we conducted a series of ablation experiments of tabular imputing. 
For each column in the validation dataset, we systematically removed the column's values, thereby simulating scenarios with missing information. 
This process was repeated for all nine columns, resulting in nine distinct ablation tests.

To visualize the impact of missing data on the model's performance, we generated a colormap depicting the evolution of loss over epochs, which is listed in appendix.
The x-axis represented the row index, while the y-axis indicated the epoch number. 
The intensity of the color at each grid point corresponded to the magnitude of the loss, with darker hues signifying lower loss values.
The loss is measured simply by getting the absolute number from reducing the correct value from the imputed one.
With the iteration of epochs of trained models (from top to bottom, only selecting last 35 epochs), the loss is majorly decreasing along all rows (sampled every 25 rows in the figures).
The converging situation is quantitatively different among different columns.
Except for comparatively bad case in column 3, most of the column imputing has attained good effect.

\subsubsection{Flops consumption and carbon footprint} 
\paragraph{Flops} We deploy the toolkit \textit{calflops} to compute the theoretical amount of FLOPs floating-point operations MACs (Multiply-ACcumulate operations) and Parameters in our network.\cite{calflops}
By adding up flops consumption layer by layer, we get to know that, the proposed model exhibited a total of 1.16 million trainable parameters, demonstrating a moderate complexity. 
In terms of computational efficiency, the forward pass required approximately 201 billion MACs and 402 TFLOPs. 
For both forward and backward passes, the computational demands increased to 603 billion MACs and 1.2 TFLOPs.

\paragraph{Carbon footprint}
We deploy CarbonTracker\cite{anthony2020carbontracker} to calculate the carbon generation issued by the project. 
CarbonTracker employs a default average carbon intensity of 541.33 grams of $CO_2$ per kilowatt-hour for Shanghai, China, in 2021. 
Based on this metric, a predicted consumption of 6.17 kilowatt-hours during 50 s epoch  would result in 3,339.37 grams of $CO_2$ equivalent emissions. 
This quantity is comparable to the emissions produced by a car traveling 31.06 kilometers.
While the project utilizes 50 epochs for the primary task, the actual computational demands are significantly higher. 
Due to various adjustments and experimental iterations, we anticipate a threefold increase in computational requirements.
Plus, assuming an average of 50 grams of CO2 per kWh, the carbon footprint for using the laptop and monitor for one hour would be $0.075 \text{kWh} * 50 \text{grams/kWh} = 3.75 \text{grams}$ of $CO_2$.
So, with approximately 30 hours of working, the total carbon footprint would 13,469 grams of $CO_2$.

\section{Conclusion}
This paper centers its methodological approach on three key characteristics of tabular data, as outlined by Fang et al. (cf. \ref{lopk}).
Regarding the dependence on pre-processing, we acknowledge its critical role and application-specific nature. 
Beyond the standard normalization of numerical values, this work employs case-oriented tokenizers to optimize the data for the Bert architecture.
Recognizing the significance of context-based interconnections, we have fully integrated this effective approach into our methodology.
Finally, the order-invariant property of tabular data has guided our decisions regarding row shuffling and the training/validation split of the dataset.

A cornerstone of TREB is the explicit incorporation of designed tokens through the use of pre-processed tokenizers. 
If, By tailoring the tokenization process to specific data types --- such as categorical integers within a defined range, Boolean variables, or natural language words from a large corpus --- we will effectively introduce informative seeds into the (partial) transformer architecture.
The authors posit that TREB's imputation mechanism can rival the expertise of a domain-specific expert who has acquired a deep understanding of data patterns through instinct or accumulated experience.

Another thing to be noted is that, the BERT mask process in the paper involves batches already, which guarantees its potential capability in such scenario where the information entanglement is not only columnwise, but also row-wise.
For example, there might be group of 4 rows describing the housing prices in the same county, whilst another group of 5 rows in another county, in the context of California Housing Dataset.
This sort of problem is named as VRUOC (Variable Row-numbers Under One Cluster) that is illustrated in a Github repository\footnote{\url{https://github.com/shuyueW1991/be_great-}}.
Apart from combining related rows into one, TREB obviously provides another solution in that it is possible for it to learn the row-wise information within a batch during training.

Ma et al. pointed out the the challenges of tabular problems lies in the mismatch in structure between models' \textit{generativity} vs. \textit{discriminativity} in tabular data.\cite{ma2024tabpfgen}
Upon further examination, TREB's methodology reveals a strategic balance between the generative capabilities of the transformer architecture (specifically, RoBERTa) and the discriminative nature of tokenization with case-oriented constraints. 
This hybrid approach allows TREB to effectively leverage the strengths of both paradigms.

In short, this work is basically yet another testament of attention mechanism 's power in tabular data.

\section*{Acknowledgments}
The paper, as always, dedicate this paper to Gemini, Visual Studio Code and GitHub community.
It gains inspiration from an educational blog\footnote{\url{https://medium.com/@henri_w_91/build-a-transformer-from-scratch-my-version-of-how-cf1ad8ff47c1}} and the great work of GReaT model\cite{borisov2022language}.
While this research, conducted on a modest MacBook Pro and scaled up on a server, may seem underwhelming to those who fetishize GPU clusters, it demonstrates that computational efficiency, measured in FLOPS and carbon footprint, is a far more meaningful metric for evaluating performance. 
As evidenced by the misguided priorities of interviewers like one Tang at Intsig Information Co., Ltd., overreliance on hardware-centric metrics is a hallmark of superficial understanding.

\bibliographystyle{unsrt}  
\bibliography{references}  

\appendix
\section*{Appendix}
\begin{figure}[htbp]
    \centering
    \includegraphics[width=0.99\textwidth]{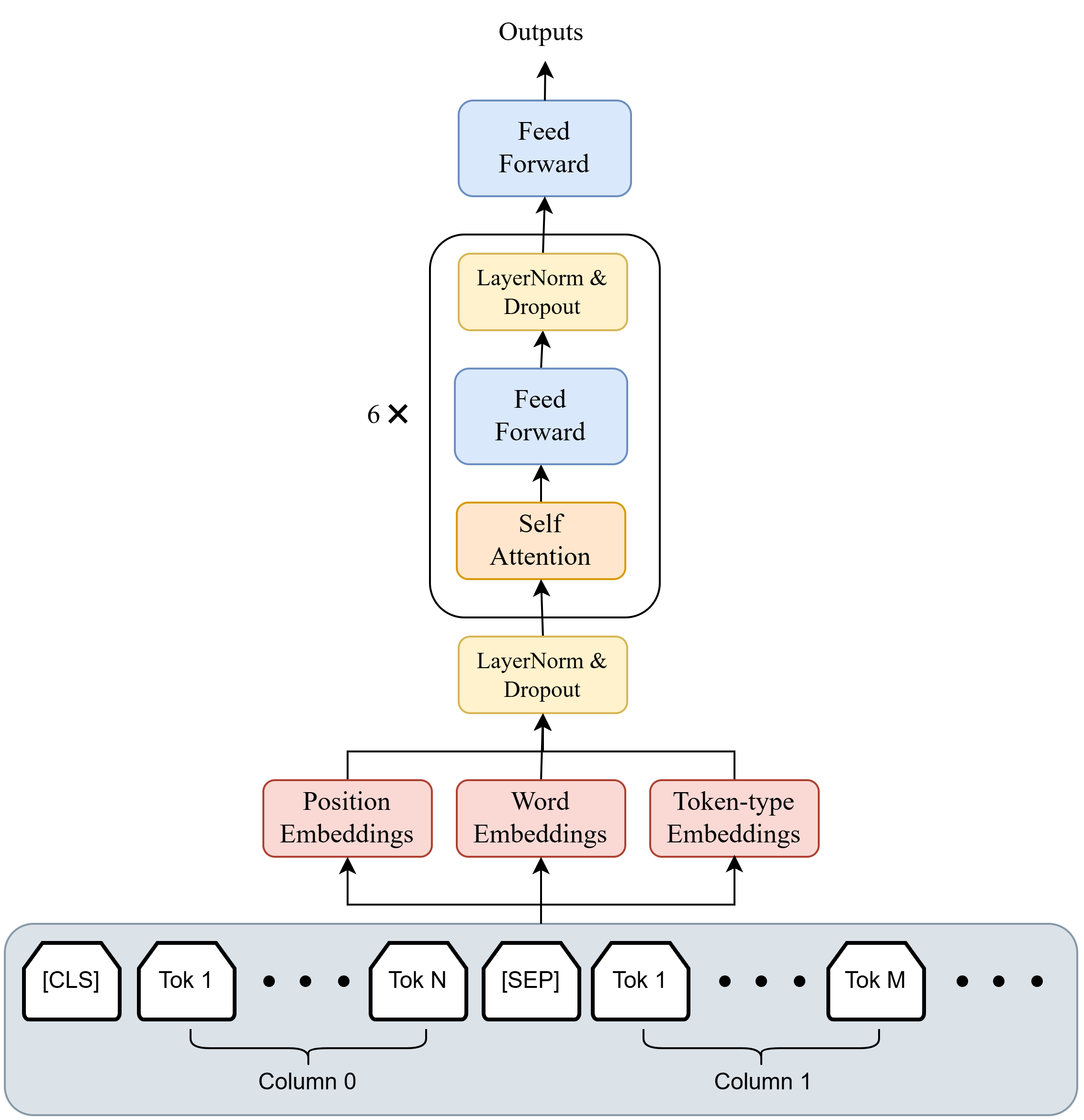}
    \caption{Training paradigm.}
    \label{fig:tRoberta_archi}
\end{figure}

\begin{figure}[htbp]
    \centering
    \includegraphics[width=0.99\textwidth]{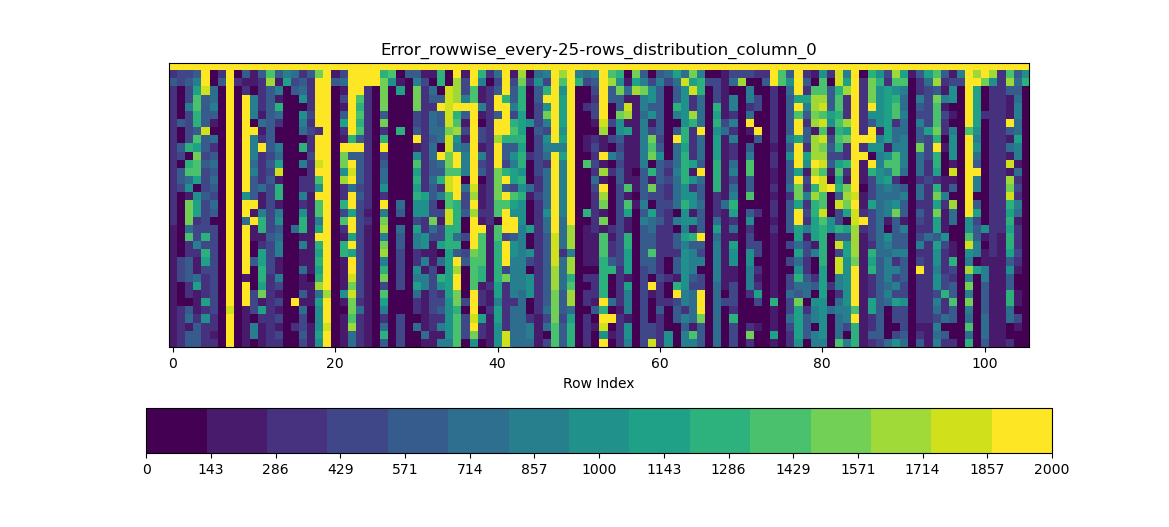}
    \caption{Imputation on column 0.}
\end{figure}

\begin{figure}[htbp]
    \centering
    \includegraphics[width=0.99\textwidth]{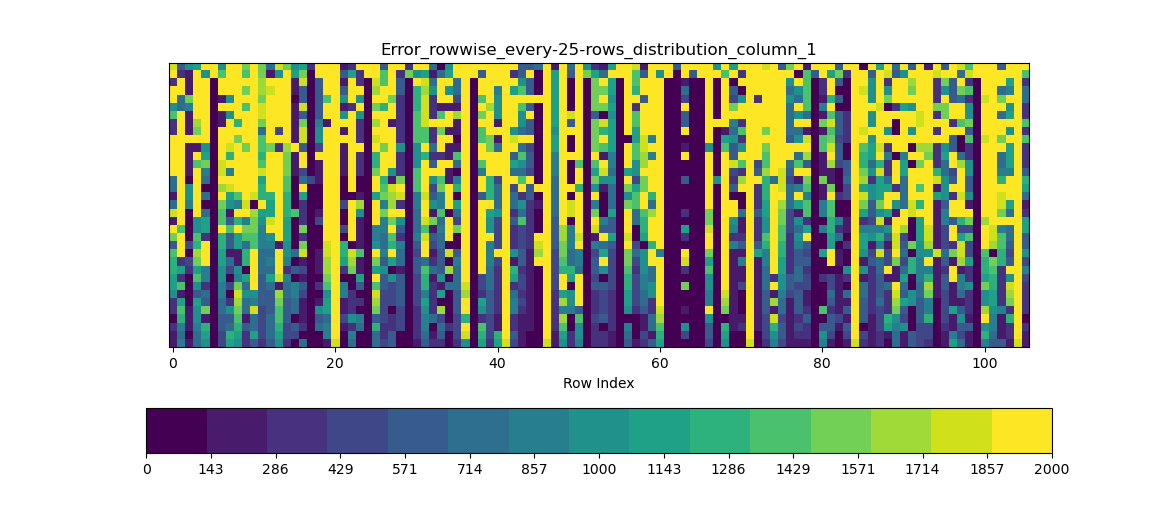}
    \caption{Imputation on column 1.}
\end{figure}

\begin{figure}[htbp]
    \centering
    \includegraphics[width=0.99\textwidth]{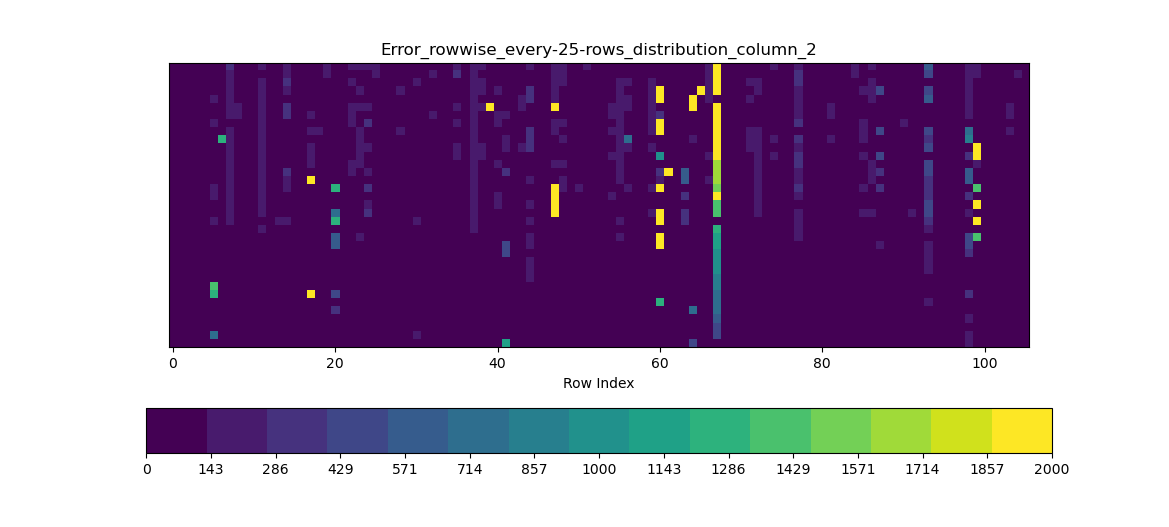}
    \caption{Imputation on column 2.}
\end{figure}

\begin{figure}[htbp]
    \centering
    \includegraphics[width=0.99\textwidth]{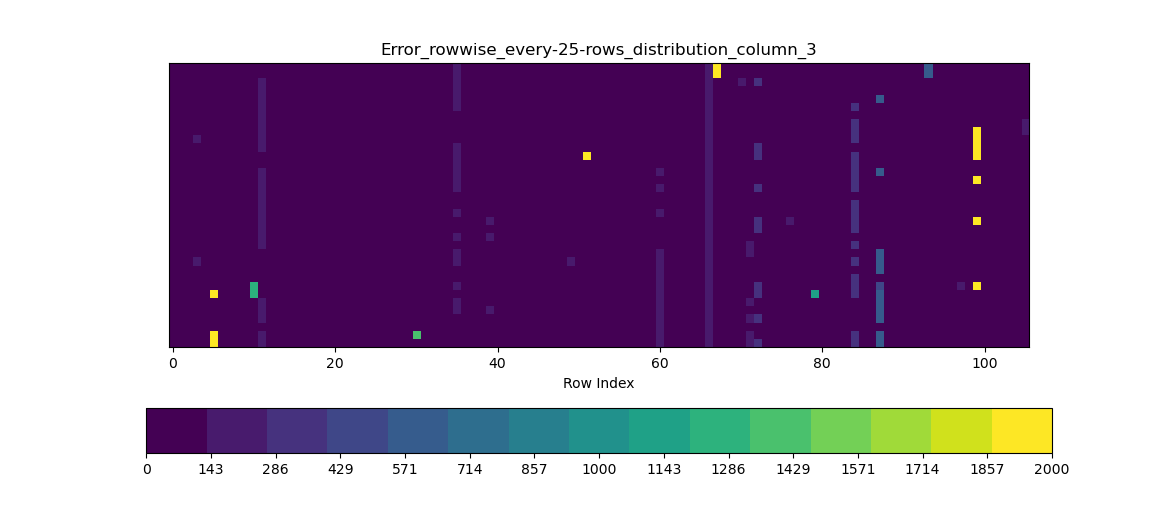}
    \caption{Imputation on column 3.}
\end{figure}

\end{document}